\documentclass[preprint,12pt,authoryear]{elsarticle}
\makeatletter
\def\ps@pprintTitle{%
 \let\@oddhead\@empty
 \let\@evenhead\@empty
 \let\@oddfoot\@empty
 \let\@evenfoot\@empty
}
\makeatother
\usepackage{amssymb}
\usepackage{xcolor}
 \usepackage{amsmath}
\usepackage[T1]{fontenc}
 \newtheorem{defn}{Definition}[section]
\usepackage{enumitem}
\usepackage{subcaption}
\usepackage{bm}
\usepackage{hyperref}
\usepackage{multirow}
\usepackage{url}
\usepackage{float}
\usepackage{afterpage}
\usepackage{tabularx}
\usepackage{geometry}
 \usepackage{booktabs}
 \usepackage{placeins}
 \usepackage{ulem}
\usepackage{float}
\usepackage{xcolor}   % For coloring text
\usepackage{hyperref} % For hyperlinking citations
\hypersetup{
    colorlinks=true,
    citecolor=blue,  % Change citation color to blue
    linkcolor=black,
    urlcolor=cyan
}
\begin{document}
\begin{frontmatter}

\title{Leveraging Advanced Machine Learning  to Predict Turbulence Dynamics from Temperature Observations at an Experimental Prescribed Fire}

\author[1]{Dipak Dulal\corref{cor2}}
\cortext[cor2]{Corresponding author}
\ead{dipak.dulal@enmu.edu}
\author[2]{Joseph J. Charney}
\author[3]{Michael R.  Gallagher}
\author[4]{Pitambar Acharya}
\author[4]{Carmeliza Navasca}
\author[5]{Nicholas S. Skowronski}

\address[1]{Department of Mathematical Science, Eastern New Mexico University, , Portales , 88130, NM, USA}
\address[2]{USDA Forest Service, Northern Research Station-2601 Coolidge Road, Suite 203, Lansing, 48910, MI, USA}
\address[3]{USDA Forest Service, Northern Research Station-501 4 Mile Road, New Lisbon, 08064, NJ, USA}
\address[4]{Department of Mathematical Science, University of Alabama at Birmingham, Birmingham, 35205, AL, USA}
\address[5]{USDA Forest Service, Northern Research Station-180 Canfield Street Morgantown, 26505, WV, USA}

\begin{abstract}
This study explores the potential for predicting turbulent kinetic energy (TKE) from more readily acquired temperature data using temperature profiles and turbulence data collected concurrently at 10 Hz during a small experimental prescribed burn in the New Jersey Pine Barrens. Machine learning models, including Deep Neural Networks, Random Forest Regressor, Gradient Boosting, and Gaussian Process Regressor, were employed to assess the potential to predict TKE from temperature perturbations and explore temporal and spatial dynamics of correlations. Data visualization and correlation analyses revealed patterns and relationships between thermocouple temperatures and TKE, providing insight into the underlying dynamics. More accurate predictions of TKE were achieved by employing various machine learning models despite a weak correlation between the predictors and the target variable. The results demonstrate significant success, particularly from regression models, in accurately predicting the TKE. The findings of this study demonstrate a novel numerical approach to identifying new relationships between temperature and airflow processes in and around the fire environment. These relationships can help refine our understanding of combustion environment processes and the coupling and decoupling of fire environment processes necessary for improving fire operations strategy and fire and smoke model predictions. The findings of this study additionally highlight the valuable role of machine learning techniques in analyzing the complex large datasets of the fire environments, showcasing their potential to advance fire research and management practices. 
\end{abstract}

%%Graphical abstract
%\begin{graphicalabstract}
%\includegraphics{grabs}
%\end{graphicalabstract}

%%Research highlights
%\begin{highlights}
%\item Research highlight 1
%\item Research highlight 2
%\end{highlights}

\begin{keyword}
%% keywords here, in the form: keyword \sep keyword, up to a maximum of 6 keywords
Fire Behavior\sep Machine Learning   \sep Thermocouple \sep Turbulence \sep Wildland Fire

%% PACS codes here, in the form: \PACS code \sep code

%% MSC codes here, in the form: \MSC code \sep code
%% or \MSC[2008] code \sep code (2000 is the default)

\end{keyword}

\end{frontmatter}

%\tableofcontents

%% \linenumbers

%% main text

\section{Introduction}
\label{introduction}

Wildland fire is a natural and essential ecological process. However, unplanned ignitions that produce wildfires have become increasingly consequential in life, property, public health, and natural resources losses \citep{Neary_2005}. Over the past decade, wildfires have intensified, with hazardous fuel conditions, extreme weather, and ignitions intersecting more frequently to produce dangerous and challenging fires with far-reaching smoke impacts. Fire behavior, smoke emission, and transport models are used to improve the prioritization and planning of safe and effective hazard reduction treatments and as a decision support tool for suppressing wildfires. However, fundamental gaps in understanding the scaling and coupling between fire and the local atmospheric processes limit the accuracy and precision of these model predictions.

Field experiments of physical fire behavior processes, such as those of RX-Cadre \citep{hiers2009overview}, the New Jersey Pine Barrens, and Camp Swift \citep{silas1}, have helped improve a fundamental understanding of the scaling and coupling of the turbulence at and around the fireline. Clark et. al \citep{fire7090330} identified a moderate correlation between the utilization of surface and understory fuels and variables such as air temperature, wind velocity, and turbulence kinetic energy (TKE) in buoyant plumes. Additionally, they conducted a quantitative analysis of various trade-offs associated with fire behavior and the turbulent dispersion of smoke during effective fuel reduction practices. Seitz et al. showed that fire-induced turbulence and heat fluxes vary greatly across even small burn areas, highlighting the need for high-resolution (1-2 m) models to capture the complex, accurately localized interactions between fire and atmosphere. The recent study by Lin et al.   \citep{egusphere-2024-2252} analyzed the effects of  background wind speed and fuel load on fire-induced wind response, turbulence characteristics revealing distinct patterns of plume tilt, smoldering, and TKE behavior under varying fire intensity conditions, and found that increased background wind speeds reduce spatial heterogeneity in fire-induced turbulence. In contrast, higher fuel loads prolong atmospheric disturbances, with both factors altering dominant turbulence components, findings critical for modeling wildfire behavior and improving early warning systems.

High-resolution time series data constitute the core of fireline field studies, offering critical insights into the intricacies of fire behavior processes and the precision of measurement techniques. The study by K. S. Shannon \citep{shannon} examines the factors influencing thermocouple accuracy in wildland fire measurements, including thermocouple type, bead size, wire size, measurement location, and environmental conditions, and provides recommendations for optimizing sensor design to improve accuracy. Hubert Luce et al.\citep{hubert} proposed and validated an analytical expression for the turbulence kinetic energy dissipation rate and the temperature structure function parameter in the convective boundary layer. Their model incorporates the Deardorff countergradient approach to account for nonlocal heat transport. In situ measurements were found to align more closely with observations than large-eddy simulation results. Volker Wulfmeyer et al.\citep{volker} have demonstrated that the concurrent use of Doppler, temperature, and water-vapor lidars can yield comprehensive profiles of molecular degradation rates and the dissipation of turbulent kinetic energy within the convective boundary layer. This was achieved by analyzing temporal autocovariance functions of wind, temperature, and moisture fluctuations. Additionally, the study by Tie Wei \citep{wei} investigates the scaling of turbulent kinetic energy and temperature variance production in a differentially heated vertical channel. The research derived and validated identity equations for global integrals of shear- and buoyancy-produced turbulent kinetic energy and temperature variance production. Moreover, it established scaling laws for these quantities at high Rayleigh and Grashof numbers, providing valuable insights into such systems' heat and energy transfer dynamics.

% Wildfires constitute intricate phenomena governed by a multitude of environmental influences, thereby requiring a comprehensive understanding of their behavioral dynamics and interaction with atmospheric components. Clark et al. \citep{atmos11030242} identified a moderate correlation between the utilization of surface and understory fuels and variables such as air temperature, wind velocity, and turbulence kinetic energy (TKE) in buoyant plumes. Additionally, they conducted a quantitative analysis of various trade-offs associated with fire behavior and the turbulent dispersion of smoke during effective fuel reduction practices. Seitz et al. \citep{acp-24-1119-2024} showed that fire-induced turbulence and heat fluxes vary greatly across even small burn areas, highlighting the need for high-resolution (1–2 m) models to capture the complex, accurately localized interactions between fire and atmosphere.
% The recent study by \citep{egusphere-2024-2252} analyzed the effects of background wind speed and fuel load on fire-induced wind response, turbulence characteristics revealing distinct patterns of plume tilt, smoldering, and TKE behavior under varying fire intensity conditions, and found that increased background wind speeds reduce spatial heterogeneity in fire-induced turbulence. In contrast, higher fuel loads prolong atmospheric disturbances, with both factors altering dominant turbulence components, findings critical for modeling wildfire behavior and improving early warning systems.
In more recent years, technologies such as deep neural networks, artificial intelligence, and machine learning have been used to to more successfully analyze fire detection and burned areas from large datasets;but have yet to be used to explore large sets of fireline data. Henintsoa S. Andrianarivony et al.\citep{Andrianarivoly} review the application of machine learning, especially deep learning techniques like convolutional and recurrent neural networks, for improving wildfire spread prediction by overcoming the limitations of traditional models and leveraging the spatiotemporal data characteristics of wildfires, ultimately aiming to enhance wildfire management and mitigation strategies. Classification and regression models have successfully predicted wildfires from remote sensing data, fire incident records, and explanatory variables such as vegetation, weather, and drought \citep{machine_Remote1,machine_Remote2}. DeCastro et al.\citep{fuel} utilized synthetic aperture radar of band C, multispectral imagery, and tree mortality survey data to successfully estimate wildland fuel data by implementing the Random Forest model. The combination of satellite spatiotemporal data and transport models has been reliable in predicting $PM_{2.5}$ concentration in large fire events \citep{machine}. 

Machine learning advancements have enhanced wildfire prediction and analysis by using diverse models and datasets for improved accuracy. Mapulane Makhaba and Simon L Winberg \citep{reinforcement}combined reinforcement \citep{reinforcement2} and supervised learning in the form of a Long-term Recurrent Convolutional Network (LRCN) to predict the spread of a large fire from its ignition point to the surrounding areas. An auto-encoder neural network (unsupervised machine learning) and coupled spatio-temporal auto-encoder (CSTAE) model have been implemented to train Sentinel 1 ground range detection (GRD) data to detect forest fires 
 \citep{unsupervised}.Bjanes et al. \citep{BJANES} proposed the use of ensemble machine learning models alongside two deep learning models to analyze anthropogenic wildfire data from Chile for the period of 2013 - 2019, with the aim of predicting wildfire susceptibility maps. They employed satellite imagery with a resolution of 25 × 25 pixels, incorporating 15 × 15 fire-influencing variables, including elevation, slope, maximum temperature, wind speed, and precipitation. Majumder et al.\citep{MAJUMDER2025102956} developed the RxFire optimization engine to identify optimal prescribed fire opportunities, using a Bayesian hierarchical model for calibrated daily weather forecasts and uncertainty estimates based on 2015-2021 data from the National Digital Forecasting Database (NDFD) and GriddMET.

\begin{figure*}[t] % Use figure* for spanning both columns
    \centering
    \includegraphics[width=\textwidth]{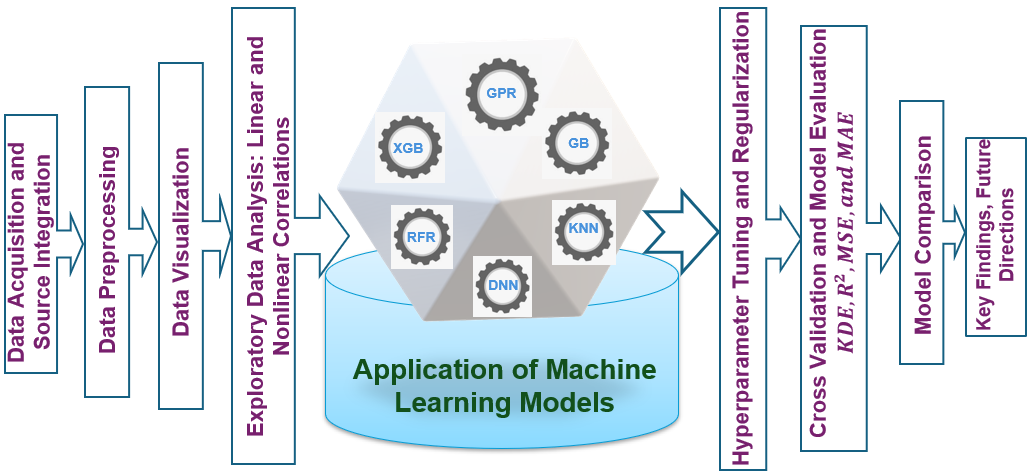} % Adjust file name and format
    \caption{Workflow of Machine Learning Model Application and Evaluation}
    \label{fig:wide_image}
\end{figure*}

The objective of this project is to employ advanced machine learning algorithms to effectively identify relationships between temperature perturbations and turbulence above a low-intensity wildland fire. By examining the correlation between thermocouple temperature measurement and TKE, this research aims to provide valuable information on the coupling and scaling of fire environment turbulence as an approach to resolving important fire behavior measurement and prediction challenges.We implemented our models on the novel data gathered during the outdoor fire experiments. See the data section (\ref{data1}).
 Figure (\ref{fig:wide_image}) represents the architecture of our entire project.

\section{Materials and Methods}

\subsection{Definitions and Preliminaries}
\label{defn}
\begin{defn} (Turbulent Kinetic Energy (TKE)) \label{tk}: TKE is a measure of the intensity of turbulence in a fluid flow \citep{whisenhant2003turbulence}. It represents the mean kinetic energy per unit mass associated with the fluctuating or chaotic motion of fluid particles, often referred to as eddies, in a turbulent flow. TKE is quantified by the mean square (MS) of the velocity fluctuations in the flow, which are the differences between the instantaneous and average velocities in each direction.
Turbulent Kinetic Energy per unit mass is defined as \citep{TKE, tkedf} 
\begin{equation}
 TKE =  \frac{1}{2}(U'^2 + V'^2 + W'^2)
\end{equation}
where $U'$, $V'$, and $W'$ represent the velocity fluctuations of $U (m/s)$, $V (m/s)$, and $W (m/s)$ in the x-, y-, and z-directions, respectively, which are deviated from their means. 

Understanding the dynamics of fire-induced turbulence is essential in researching wildland fires. Turbulent eddies play a crucial role in heat transfer, the dispersion of firebrands, and the quantification of energy contained in them. These studies are possible through TKE measurements. By modeling fire-atmosphere interactions with the help of TKE measurements, we can improve our predictions of the behavior of fire propagation under different environmental conditions.
\end{defn}

\begin{defn}\label{th} (Thermocouple Temperature): 
Thermocouple temperature is a type of temperature measurement that utilizes a thermocouple. A thermocouple is a temperature sensor made from two dissimilar metal wires joined at one end.

Due to their durability, wide temperature ranges, and fast response time, thermocouples are widely used in various industrial and scientific applications. More importantly, they are widely used in wildland fires due to their ability to measure high temperatures accurately and withstand harsh conditions. Thermocouples are useful in fire behavior analysis, fuel consumption studies, research, and experimental burns. 
\end{defn}
\begin{defn}
    (Ridge Regularization ($\ell_2$ Norm)): The Ridge regularization, also known as Tikhonov regularization, is a statistical technique that  adds a penalty to the objective function of an optimization problem to prevent overfitting in linear regression models by reducing model complexity through shrinking the coefficients. It is particularly useful when dealing with multicollinearity among independent variables, leading to unstable models with high variance. Mathematically, the Ridge penalty term ($\ell_2$ norm) can be defined as:

\[
\|\omega\|_2^2 = \sum_{i=1}^n \omega_i^2
\]

where $\omega = (\omega_1, \omega_2, \ldots, \omega_n) \in \mathbb{R}^n$ represents the coefficients vector in a regression model.
\end{defn}
\begin{defn}
    (Lasso Regularization ($\ell_1$ Norm): The Lasso regularization also adds a penalty to the objective function to reduce model complexity and overfitting, shrinking the coefficients and setting up some of them to zero. It is widely used for sparse optimization problems. It is defined as  
    \[
\|\omega\|_1 = \sum_{i=1}^n |\omega_i|
\]

where $\omega = (\omega_1, \omega_2, \ldots, \omega_n) \in \mathbb{R}^n$ represents the coefficients vector in a regression model.
\end{defn}
\begin{defn}
  (Elastic Net Regularization): Elastic Net is a linear regression methodology that integrates $\ell_1$ and $\ell_2$ regularization techniques. It effectively addresses multicollinearity and facilitates feature selection. This approach is particularly advantageous when managing datasets with highly correlated features or when the number of features surpasses the number of observations. Elastic Net mitigates some of the limitations associated with Lasso and Ridge regression by equilibrating their strengths and weaknesses. The regularization term in Elastic Net is expressed as:
  \begin{equation*}
      \lambda_1 \| \omega \|_1 + \lambda_2 \|\omega\|_2
  \end{equation*}
  If we explore more mathematically \citep{Horn_Johnson_1985}, we have an equivalent relationship
  \begin{equation*}
       \| \omega \|_2 \leq  \|\omega\|_1 \leq  \sqrt{n} \| \omega \|_2   
 \end{equation*}
 It follows that
 \begin{equation*}
    \lambda_1  \|\omega\|_1  + \lambda_2 \| \omega \|_2 \leq  (\lambda_1 + \lambda_2) \|\omega\|_1 
  \end{equation*}
  and
  \begin{equation*}
    \left( \lambda_1 +  \frac{\lambda_2}{\sqrt{n}} \right) \|\omega\|_1 \leq \lambda_1  \| \omega \|_1 + \lambda_2  \| \omega \|_2 
 \end{equation*}
 Thus,
 \begin{equation*}
    \left( \lambda_1 +  \frac{\lambda_2}{\sqrt{n}} \right) \|\omega\|_1 \leq \lambda_1  \| \omega \|_1 + \lambda_2  \| \omega \|_2 \leq (\lambda_1  +\lambda_2) \| \omega \|_1 
 \end{equation*}
 The above mathematical equation indicates that conducting a detailed grid search to optimally tune lasso and ridge regression can yield results that are comparable to those of the elastic net approach. However, the elastic net method inherently combines the benefits of both techniques, promoting sparsity while minimizing instability during the optimization process. This creates a balance between smoothness and sparsity, leading to more efficient and stable updates, which ultimately saves time and computational resources during simulations.

  In this context, $\lambda_1$ and $\lambda_2$ are hyperparameters governing the contribution of $\ell_1$ and $\ell_2$ norms.
\end{defn}
\begin{defn}
    (Pearson's Correlation Coefficient) Pearson's correlation  between two variables \(X\) and \(Y\), and is defined as:
\[
r = \frac{\sum\limits_{i=1}^N (X_i - \overline{X})(Y_i - \overline{Y})}{\sqrt{\sum\limits_{i=1}^N (X_i - \overline{X})^2 \sum\limits_{i=1}^N (Y_i - \overline{Y})^2}}
\]
$X_i$ and $Y_i$ represent individual sample points, while $\overline{X}$ and $\overline{Y}$ represent mean values for X and Y variables, respectively, and n is the number of data points. 

Pearson's correlation is commonly used in machine learning to measure linear relationships between features and targets. It helps identify highly correlated features for model training and removes redundant features to improve model performance and interpretability.
\end{defn}
\begin{defn}
    (Spearman's Rank Correlation Coefficient)
    Spearman's rank correlation coefficient, or Spearman's rho, assesses how well the relationship between two variables can be described using a monotonic function. It is calculated as:
\[
\rho = 1 - \frac{6 \sum\limits_{i=1}^N d_i^2}{N(N^2 - 1)}
\]
where \(d_i\) is the difference between the ranks of corresponding values \(X_i\) and \(Y_i\), and \(N\) is the number of observations.

Spearman's correlation is used when data does not meet the assumptions of Pearson's correlation. It can handle non-linear relationships and ordinal variables. It is often used in machine learning for exploratory data analysis and feature selection.

\end{defn}
\begin{defn}\label{sonic}(Sonic Temperature) \citep{thermo}: 
Sonic temperature is a parameter that is calculated using the speed of sound through air. This parameter is affected by the temperature of the air. Sonic temperature provides an accurate and reliable method of measuring the atmosphere's temperature without physical contact with the air being measured. 

The calculation of sonic temperature is based on the equation for the speed of sound in air \citep{thermo}, which is given by $c = \sqrt{\gamma R T/M}$. In this equation, $c$ represents the speed of sound, $\gamma$ is the ratio of specific heats, $R$ is the universal gas constant, $T$ is the absolute temperature in Kelvin, and $M$ is the molar mass of the air. Sonic temperature is instrumental in studying turbulent flow, where traditional temperature-measuring devices such as thermocouples or RTDs cannot accurately capture rapid temperature fluctuations. In wildland fire research and management, sonic temperature provides unique advantages, especially in comprehending fire-atmosphere interactions, fire behavior, and the impacts of fire on the environment.
\end{defn}

\subsection{Methodology}

\subsection{Problem Formulation}

Let our dataset be represented as
\begin{equation}
  \mathcal{D} = \{(\mathbf{x}^{(i)}, \mathbf{y}^{(i)})\}_{i=1}^{N} 
\end{equation} where $\mathbf{x}^{(i)} \in \mathbb{R}^8$ such that
\[
\mathbf{x}^{(i)} = [T1^{(i)}, T2^{(i)},T3^{(i)}, T4^{(i)},T5^{(i)},T6^{(i)},T7^{(i)},\text{sonic}\_T^{(i)} ]
\]
 And \(\mathbf{y}^{(i)} \in \mathbb{R}\) is the corresponding output,i.e. TKE.

Our objective is to train a machine learning model \(f(\mathbf{x}; \boldsymbol{\theta})\) parametrized by \(\boldsymbol{\theta}\) to approximate the function \(f: \mathbb{R}^8 \to \mathbb{R}\) that maps input vectors to TKE values. The objection function for our optimization problem is formulated as follows:
\begin{equation}
\boldsymbol{\theta}^* = \arg\min_{\boldsymbol{\theta}} \left( \sum_{i=1}^{N} L(\boldsymbol{y}^{(i)}, f(\mathbf{x}^{(i)}; \boldsymbol{\theta})) + \lambda R(\boldsymbol{\theta}) \right)
\end{equation}
where \(L\) is a loss function $R$ is a regularization term, and \(\lambda > 0\) is the regularization strength.
\subsection{Model Implementation and Description}
This section clarifies the mathematical insights, their significance to our objectives, and the datasets used for each implemented machine learning model. We have chosen to focus on the models that exhibited the best overall performance metrics.

\subsubsection{Deep Neural Network (DNN)}
\begin{figure}[H]
    \centering  \includegraphics[scale= 0.22]{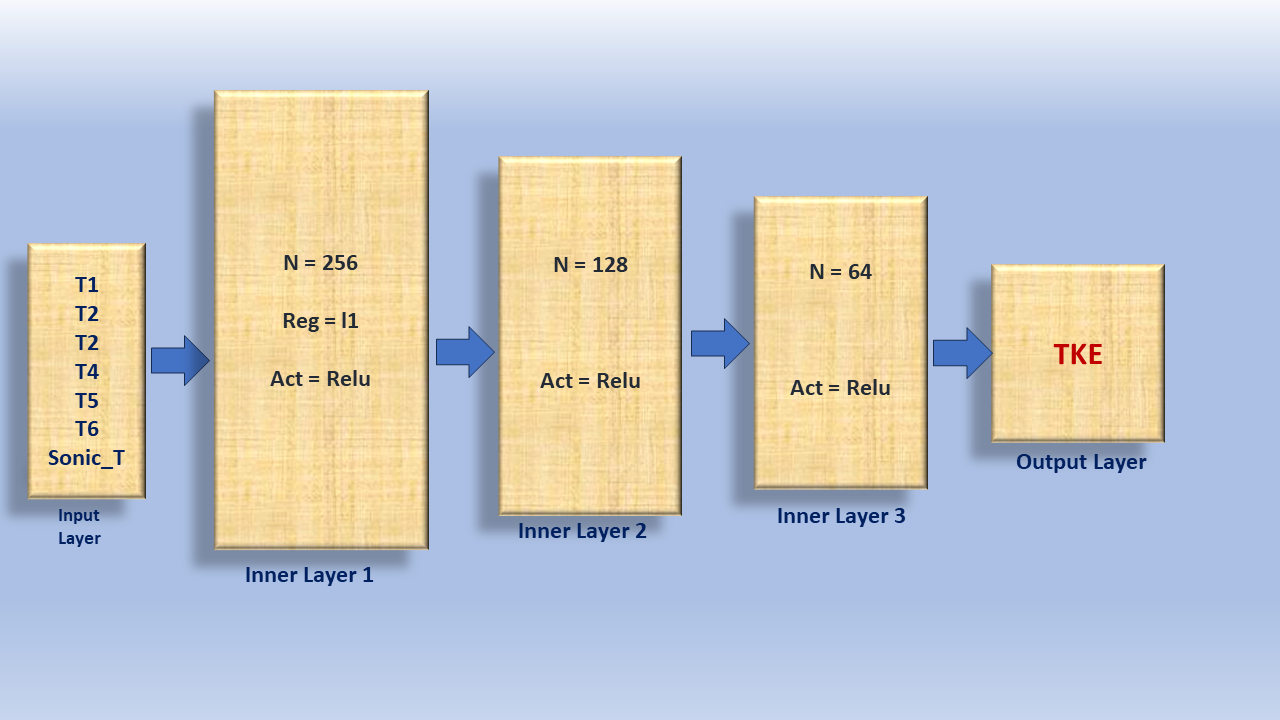}
    \caption{DNN Architecture}
    \label{fig:enter-dnn}
\end{figure}

Deep Neural Networks (DNNs) are complex models used in machine learning to solve a wide range of problems, from image recognition to natural language processing. The deep neural network (DNN) comprises an input layer, three hidden layers, and an output layer. Each layer plays a crucial role in processing and transforming the input data to achieve the desired output. The layers are interconnected through weights and biases, optimized to minimize the objective function \citep{deep2}. This optimization process, often called backpropagation, allows the network to learn from its errors and improve its performance over time. DNN's ability to automatically learn hierarchical features from raw data has revolutionized many areas of artificial intelligence and machine learning. It's worth noting that the specific architecture of a DNN can vary depending on the task at hand. For instance, Convolutional Neural Networks (CNNs) are specialized DNNs designed for image processing tasks, while Recurrent Neural Networks (RNNs) are better suited for sequential data like text or time series
Figure (\ref{fig:enter-dnn}) reflects our DNN algorithm structure.

\subsubsection{Random Forest Regressor (RF)}
Random Forest aggregates the predictions of multiple decision trees, offering an ensemble prediction:

\begin{equation}
    \hat{y} = \frac{1}{M} \sum_{m=1}^{M} T_m(\bm{T})
\end{equation}
where \( M \) is the number of trees and $T_m(T)$  represents the prediction from the m-th tree. By training each tree on a bootstrapped subset of the data and randomly selecting features at each split, RF introduces diversity, reducing variance and mitigating overfitting. Additionally, it provides feature importance metrics, aiding in feature selection. While computationally intensive for large datasets, RF is robust, handles high-dimensional data effectively, and is widely used in classification and regression tasks.

\subsubsection{K-Nearest Neighbors (k-NN) Regressor}

The Euclidean distance, also known as the $\ell_2$ norm, calculates the square root of the sum of the squared differences between matching components of two vectors. This measure represents the shortest path, or straight-line distance, between two points within a Euclidean space.
\[ D(x, y) = \sqrt{\sum_{i=1}^{d} (x_i - y_i)^2} \]

k-NN predicts TKE by averaging the values of the \( k \)-nearest training samples.  
Given a dataset \( \mathcal{D} \) and \( y^{(i)} \in \mathbb{R} \) as the corresponding TKE output, the k-NN regressor estimates the TKE as:

\begin{equation}
    \hat{y} = \frac{1}{k} \sum_{i \in \mathcal{N}_k(\bm{T_*})} y^{(i)}
\end{equation}

where \( \mathcal{N}_k(\bm{T_*}) \) denotes the set of \( k \)-nearest neighbors and $y^{(i)}$ is the TKE value for the $i^{\text{th}}$ neighbor. 
The choice of $k$ and the distance metric significantly impact prediction accuracy. Smaller values of $k$ make the model sensitive to noise, while larger values may smooth out local variations. 
 Additionally, k-NN is non-parametric, meaning it does not assume any underlying distribution for the data.
.

\subsubsection{Gradient Boosting (GB) Regressor}
GB constructs  predictive models by sequentially adding decision trees, with the prediction at stage $T$ given by:

\begin{equation}
\hat{y}(\bm{T}) = \sum_{j=1}^{J} \rho_j T_j(\bm{T})
\end{equation}

\noindent where $J$ is the number of boosting stages, $T_j(T)$ is the $j$-th tree's output, and $\rho_j$ is the step sizes (learning rate). The process starts with an initial prediction, then iteratively computes residuals, fits a new tree to these residuals, and updates the model by adding the scaled tree output. GB minimizes a loss function via gradient descent, computing pseudo-residuals as gradients, fitting trees to these gradients, and updating predictions, with key parameters including learning rate $\rho_j$, number of trees $J$, and tree depth.

\subsubsection{Extreme Gradient Boosting(XGB)}

XGBoost's objective integrates the number of leaves and leaf scores, further regularized by the $\ell_1$ and $\ell_2$ norms \citep{xboost}:

\begin{equation}\label{xgb}
\begin{split}
\mathcal{F}(\theta) &= \sum_{i=1}^{n} L(y^{(i)}, f({x^{(i)}}; \theta)) + \gamma T \\
&\quad +\lambda_1 \sum_{j = 1}^{T}\vert w_j\vert + \frac{1}{2}\lambda_2 \sum_{j = 1}^{T} w_j^2.
\end{split}
\end{equation}

Where $T$ denotes the quantity of leaves essential for regulating the complexity of model structures penalized with the parameter $\gamma$, and \( w_j \) denotes the scores of the leaves \citep{xboost}. The parameters $\lambda_1$ and $\lambda_2$ are regularization parameters that govern the intensity and level of contribution of the $\ell_1$ and $\ell_2$ regularization, respectively, to ensure the optimum output.

\subsubsection{Gaussian Process Regression with Regularization}

The optimization problem of the Gaussian Process Regressor (GPR) model is to maximize the log marginal likelihood (LML) of the observed data \citep{gp}. LML for our dataset $\mathcal{D} = \{(x_i, y_i)\}_{o = 1}^N$, where $x_i$ are input features and $y_i$ are output features with N observations. The corresponding objective function is given below

\begin{equation}
\begin{split}
    \mathcal{F}(\theta) &= \frac{1}{2}(y-m(X))^T(K(X,X) + \sigma_n^2I)^{-1}(y-m(X)) \\
    &\quad - \frac{1}{2} \log\vert K + \sigma_n^2I\vert - \frac{N}{2} \log(2\pi).
\end{split}
\end{equation}

where is a covariance matrix of training datasets $X$ of size N, which consists of kernel function; $k(x_i,x_j;\theta)$, where $x_i$, and $x_j$ are training feature vectors of the $i^{th}$ and $j^{th}$ observations, respectively; see \citep{gp}. The function m(X) represents the mean function of each column of input matrix X, and the term $\sigma_n^2$ acts as a form of regularization to account for the noise present in the target variable. Furthermore, $\sigma_n^2$  also helps prevent overfitting of the data and guarantees that $K + \sigma_n^2I$ remains positive definite. The parameter $\theta$ represents the hyperparameters of the kernel function, such as length scales, variance parameters, etc. The initial term $- \frac{1}{2}y^T(K + \sigma_n^2I)^{-1}y$ evaluates how well the model fits the training data, considering the model's complexity determined by the kernel. The final term $-\frac{N}{2}log(2\pi)$ introduces a bias related to the hyperparameters. We have tuned the regularizers with the parameters for the optimum possible output. Radial Basis Function (RBF) kernel \citep{gp}, WhiteKernel \citep{gp}, and RationalQuadratic \citep{gp} regularizers were implemented in our model.
\subsection{Summary of Model Parameters}
The table (\ref{model_table}) comprehensively summarizes the optimized hyperparameters employed in evaluating machine learning models within this study. The dataset was systematically partitioned into training (64\%), testing (16\%), and validation (20\%) subsets, with Standard Scaler applied for data normalization. The deep neural network (DNN) architecture consists of three hidden layers and employs the Adam optimizer, set with a learning rate of 0.0001, alongside $\ell_1$ regularization ($\lambda=0.01$) and a ReLU-softmax activation function. Our investigation was mainly directed towards understanding the mathematical implications and their penalty impact on the models concerning $\ell_1$ and $\ell_2$, as well as their intersection, in pursuit of optimal performance. A Grid-search methodology was utilized to determine the optimal learning parameter for achieving the most effective optimization outcome. Presenting this information in a clear, tabular format facilitates a deeper understanding of the model's inner workings and the decisions made during its development and optimization phases.

\begin{table*}[!htbp]
\begin{minipage}{\linewidth}
\centering
\begin{tabular}{@{}lp{13.5cm}@{}}
\hline
\hline
\textbf{Universal} & \textbf{Data Split ( Train, Test, Val.):} 64\%-16\%-20\%, \textbf{Scaling:} StandardScaler \\
\hline
\vspace{.4cm}
\textbf{ML Model} & \\

\textcolor{blue}{DNN} & \textbf{Inner Layers:} 3, \textbf{Optimizer:} Adam (learning rate =0.0001), \\ &\textbf{Loss:} MSE, \textbf{Batch:} 42, \textbf{Epochs}: 1000, \textbf{Patience:} 10, \\& \textbf{Regularization:} $l_1$ ($\lambda=0.01$),\textbf{Activation: } Relu +  softmax \\

\textcolor{blue}{RFR} & \textbf{Estimators}: 200, \textbf{Max Features}: Auto, \textbf{Depth}: 10, \textbf{Min Split:} 2,\\ & \textbf{CV}: ShuffleSplit (5, 80-20) \\

\textcolor{blue}{KNN} & \textbf{Neighbors:} 3, \textbf{Distance:} Euclidean, \textbf{Weights:} Distance \\

\textcolor{blue}{GBR} & \textbf{Rate:} 0.2, \textbf{Estimators:} 100 \\

\textcolor{blue}{GPR} & \textbf{Kernel:} RBF (length\_scale = 1) + WhiteKernel (noise\_level = 1)\\& + Rational Quadratic, \textbf{Alpha:} 0.01 \\

\textcolor{blue}{XGB} & \textbf{Estimators:} 200, \textbf{Depth}: 10, \textbf{Learning Rate:} 0.01,\\ & \textbf{Regularizer:} \textbf{$L_1$:} 1, \textbf{$L_2$:} 1.5 \\
\hline
\end{tabular}
\end{minipage}
\caption{\textbf{Fine Tuned Parameters and Hyperparameters}} 
\label{model_table}
\end{table*}  
\subsection{Evaluation Metrics}
We have meticulously employed the following three evaluation metrics to ascertain the effectiveness and resilience of our machine learning models.
\subsubsection{Coefficient of Determination ($R^2$)}The coefficient of determination, denoted as $\mathbf{R}^2$, is a statistical instrument that quantifies the extent to which a regression model accounts for the variance observed in the target variable. It is a widely employed metric for assessing model performance, particularly in the context of our ensemble learning prediction models. The $\mathbf{R}^2$ is defined as
\begin{equation}
    \mathbf{R}^2 = 1- \frac{\sum\limits_{i=1}^N (\mathbf{(y_i-\hat{y_i})^2}}{\sum\limits_{i=1}^{N} (\mathbf{(y}_{i}-\overline{y_i})^2}
\end{equation}
where the $\mathbf{\hat{y}}$ denotes our predicted TKE values, and $\mathbf{\overline{y}}$ is the mean of actual TKE values $\mathbf{y^i}$
\begin{equation*}
    \mathbf{\overline{y}}= \frac{1}{N}\sum\limits_{i=1}^N\mathbf{{y^i}}
\end{equation*}
As the value of \(R^2\) approaches 1, the model's predictions indicate perfect accuracy. Conversely, a value of \(R^2\) near 0 signifies poor performance in predicting the TKE values \(\hat{y_i}\).
\subsubsection{Kernel Density Estimation Graph (KDE)}Kernel Density Estimation \citep{kde2} is a non-parametric method employed to approximate the probability density function of a random variable. This efficacious tool has been utilized to evaluate our machine learning models, specifically to visualize the distribution of actual versus predicted TKE values across various algorithms. It aids in assessing the dispersion and bias within model predictions and identifying outliers and irregularities. We used Gaussian kernel function $\mathbf{K}$ defined as 
\begin{equation}
    \mathbf{k(x,x_i)}=\frac{1}{\sqrt{2\pi}}e^{-\frac{(\mathbf{x-x}_i)^2}{2h}}
\end{equation} for KDE to estimate density function $f(\mathbf{x})$ with random datasets $x_1, x_2, \cdots, x_N$ with bandwidth h defined as:
\begin{equation}
    f(x) = \frac{1}{Nh}\sum\limits_{i=1}^N \mathbf{K((x,x_i)}
\end{equation}
The kernel density estimation (KDE) plot demonstrates that a shape centered at zero and exhibiting symmetry indicates strong performance. Conversely, shifts to the left or right suggest the presence of bias, while a flat distribution may signal high variance.
\subsection{Mean Squared Error (MSE)} Mean Squared Error serves as a metric for assessing the performance of predictive machine learning models, emphasizing regression models. It measures the divergence of a model's predictions from the estimated TKE based on actual TKE data. Mathematically:
\begin{equation}
    MSE = \frac{1}{N}\sum\limits_{i=1}^N({\mathbf{y}_i- \hat{\mathbf{y_i}}})^2
\end{equation}
A lower MSE value indicates superior model performance. This metric demonstrates stability to gradient updates in the optimization process. Nevertheless, it penalizes large errors due to the squaring aspect.
\subsubsection{Mean Absolute Error(MAE)}
Model fit for the continuous variables is evaluated on the basis of the mean absolute error. It is the absolute mean difference between the predicted values and actual values, defined as
\begin{equation*}
    MAE = \frac{1}{N}\sum\limits_{i=1}^N \vert\mathbf{y_i- \hat{y_i}\vert}
\end{equation*}
This has a lower impact of outliers than MSE and treats all errors equally.
\section{Data Sources and Preprocession}
The dataset utilized in this research originates from a series of outdoor fire experiments that were designed to mimic operational prescribed burns fire experiments aimed at studying fire behavior, atmospheric interactions, and fuel dynamics related to combustion processes (including fuel bed characteristics, heat transfer, flame propagation, and airflow). These experiments also sought to bridge the gap between small-scale laboratory combustion studies and large-scale operational prescribed burns. Collected data encompasses various attributes, such as wind speed, temperature, and infrared imagery, providing a comprehensive view of fire dynamics. Given the dataset's complexity and multi-format nature—incorporating multi-band imagery and structured CSV files—rigorous preprocessing was necessary to ensure analytical consistency and usability. This section details the data acquisition process, preprocessing methods, and visualization techniques employed to extract meaningful insights relevant to machine learning applications.
\subsection{Data Aquisition}\label{data1}
\begin{figure*}[!htbp]
    \centering
    \subfloat[Equipment setup]{\includegraphics[width=0.30\textwidth]{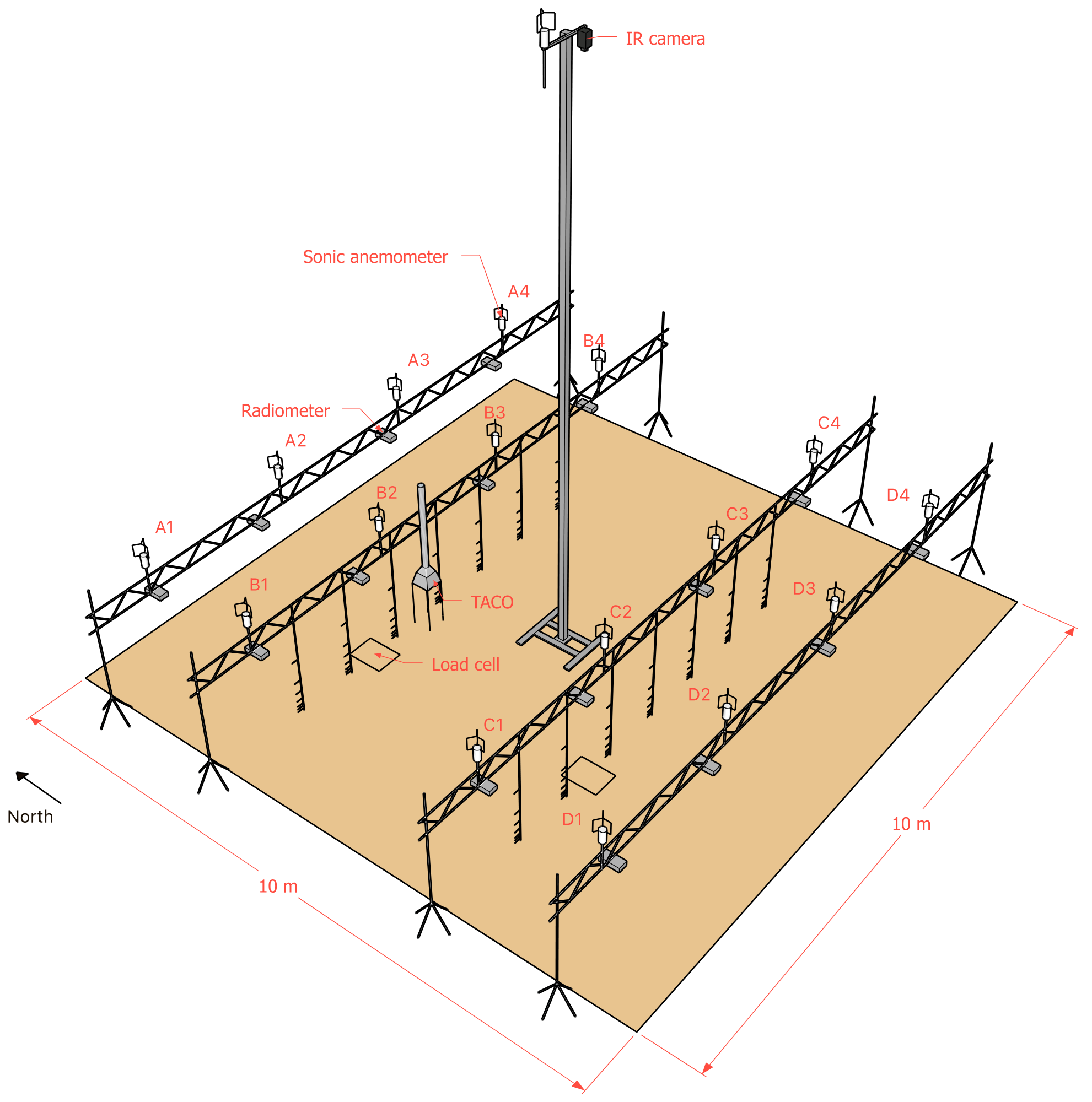}\label{fig:set-up}}
\hspace{0.02\textwidth} % Reduce horizontal gap
\subfloat[Thermocouple setup]{\includegraphics[width=0.32\textwidth]{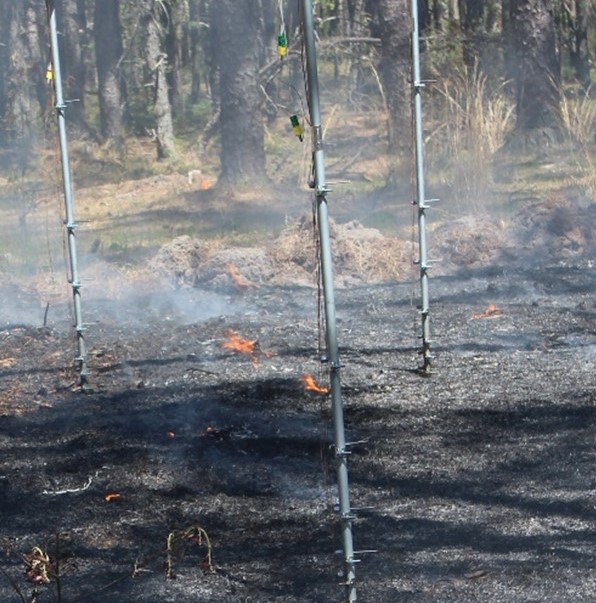}\label{fig:thermo}} \\
\subfloat[Field view]{\includegraphics[width=0.28\textwidth]{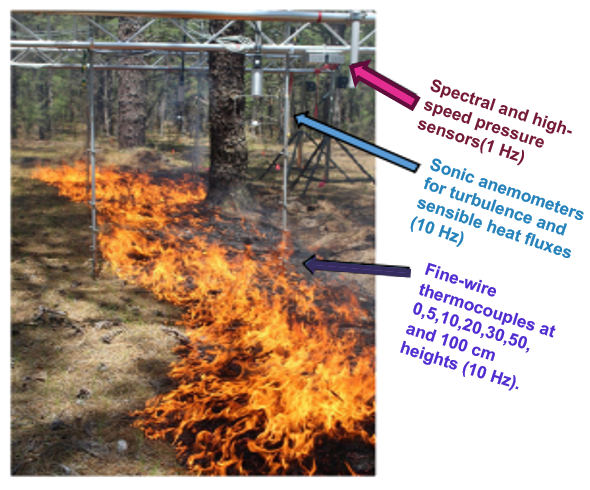}\label{fig:burn}}
\hspace{0.034\textwidth} % Reduce horizontal gap
\subfloat[Infrared images]{\includegraphics[width=0.32\textwidth]{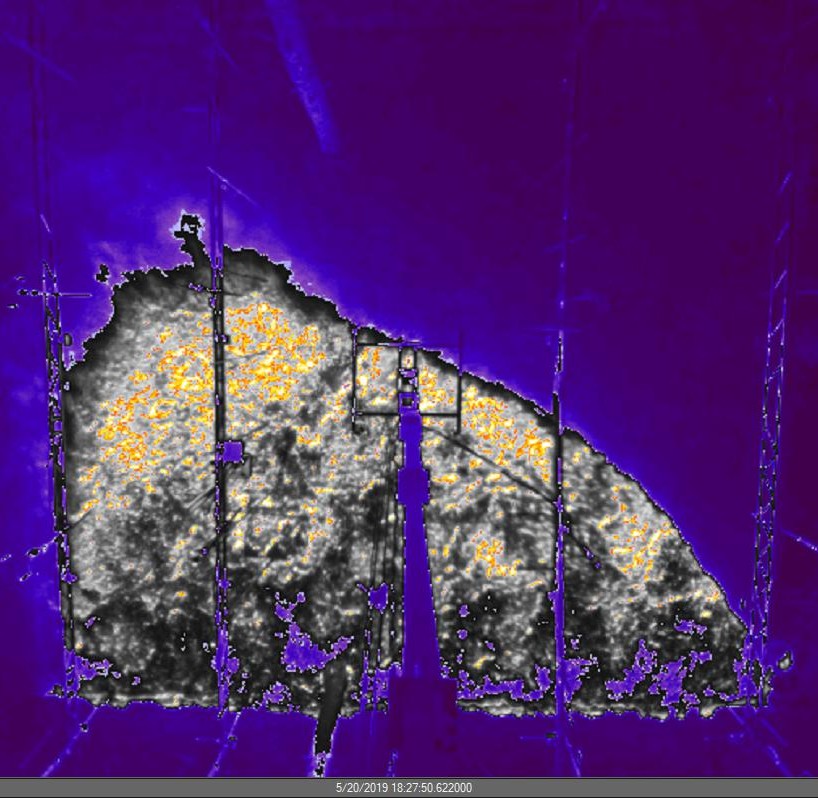}\label{fig:ir}}
    \caption{The instrument setups used for data collection and the images captured during the burn on a $10$m x $10$m field of Burn 20. Figure (a) provides a detailed overview of the equipment setup for data collection, (b) shows the status of the thermocouple setup, while (c) displays the field, and (d) presents the infrared (IR) images taken during the burn \citep{Skowronski2021}.}
    \label{fig:burning_images}
\end{figure*}

 Data were acquired from rigorously instrumented small-scale fire experiments carried out at the Silas Little Experimental Forest (SLEF) in New Lisbon, New Jersey ($39.91^o N$, $74.6^o W$, $~ 1013 Mb$), which presents a flat terrain, sandy acidic soils, oak-pine ecosystems, and a humid continental climate conducive to prescribed fire studies. All experiments in the series were carried out within 10m $\times$ 10m research plots located within the SLEF research plantation, spanning from March 2018 to June 20l9 \citep{silas1,silas2,silas3}. Figures (\ref{fig:set-up}),(\ref{fig:thermo}),(\ref{fig:burn}), and (\ref{fig:ir}) illustrate the equipment setup's design, the anemometer's placement, and images, both visible and infrared, taken during the burning period.
  During the experiments, data from multiple attributes of the fire were characterized in different formats: multi-band imagery with digital numbers, CSV files, and TIF data. However, our particular emphasis was on analyzing the sonic anemometer and thermocouple temperature data stored in CSV format. The main focus of the analysis presented here revolved around the B and C trusses, each subdivided into four distinct groups, as illustrated in Figure (\ref{fig:set-up}).\\

\subsection{Data Preprocessing and Visualization}
\begin{figure*}[ht]
    \centering
    \begin{minipage}{\textwidth}
        \centering
        \includegraphics[width = \textwidth]{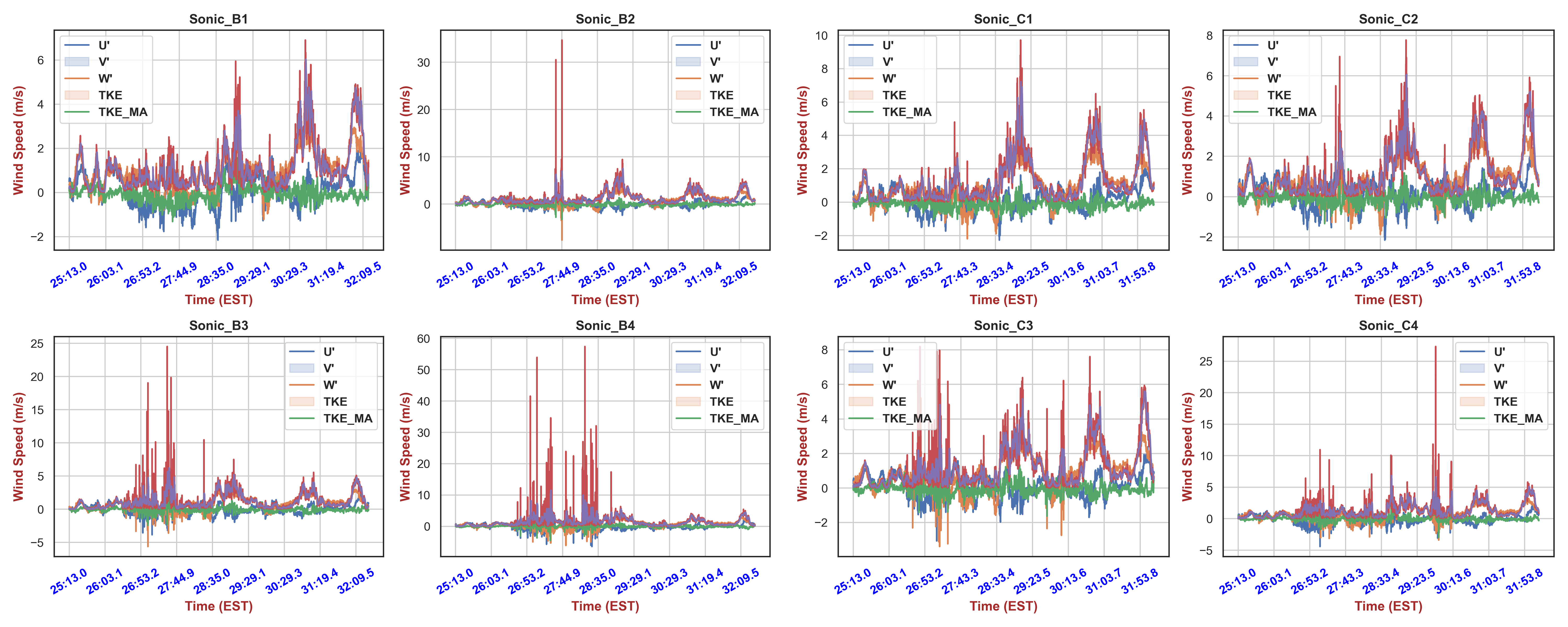}
        \caption{Wind Speeds and Computed TKE collected from B1- B4 and C1-C4 trusses of Burn 20.}
        \label{fig:sonic_speed}
    \end{minipage}

 \vspace{0.2cm}

    \begin{minipage}{\textwidth}
        \centering
        \includegraphics[width = \textwidth]{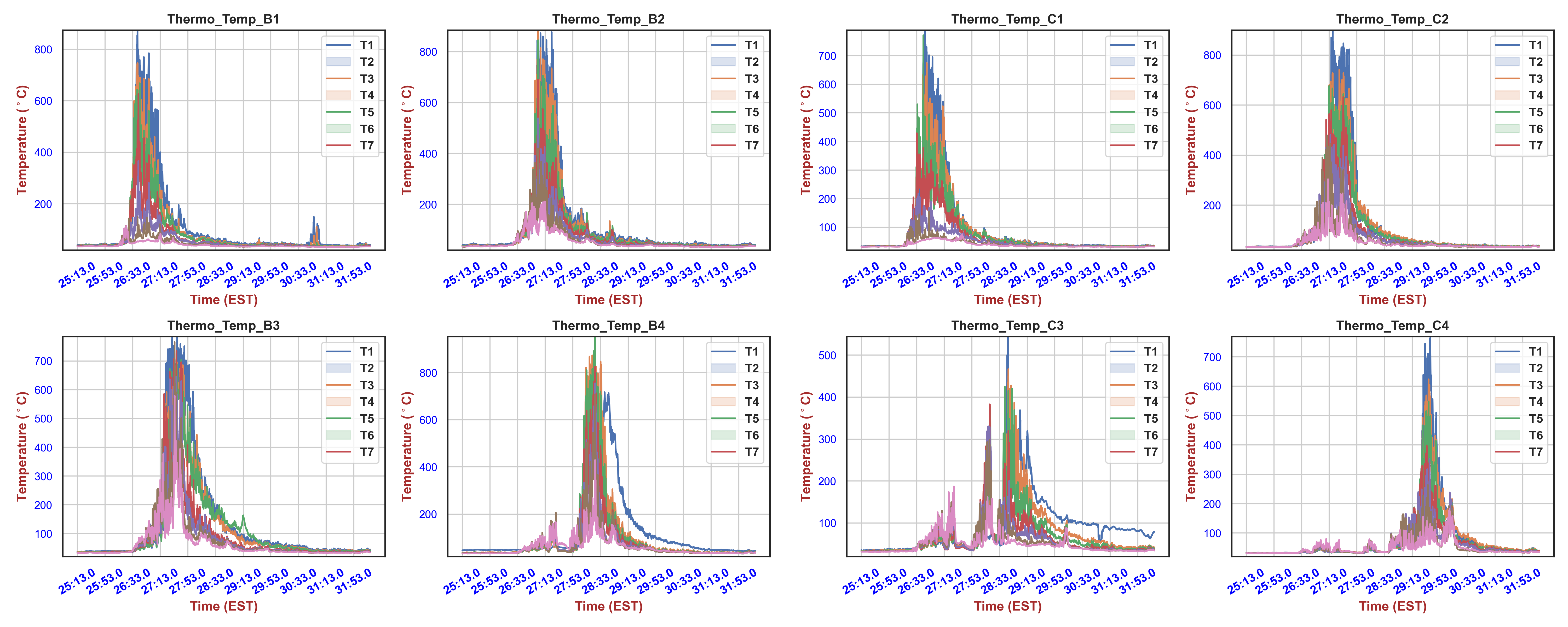}
        \caption{Thermocouple temperatures from different heights from B1- B4 and C1-C4 trusses of Burn 20. }
        \label{fig:thermocouple}
    \end{minipage}
\end{figure*}

We implemented two datasets collected from trusses B and C from Burn 20, each with four clusters: B1, B2, B3, B4, and C1, C2, C3, and C4, respectively. We divided the data into three parts: pre-burn, burn, and post-burn, based on the timestamps. Our correlation and machine learning model exploration concentrates on the burn period section. We computed our target variable TKE  as defined in the definition (\ref{tk}).
The graphical representation of the computed TKE exhibited incremental fluctuations. Consequently, we calculated the 10-point moving average of TKE, which we denote as TKE\_MA.. Figure (\ref{fig:sonic_speed}) displays the wind speeds, the TKE, and the TKE moving average ($TKE\_MA$).We focused on observing and executing our models on the anemometer temperatures($^oC)$ $T_1, T_2, T_3, T_4, T_5, T_6$, and $T_7$ from the trusses mounted at 0, 5, 10, 20, 30,50, and 100 cm above the fuel beds. We truncated the temperatures between $-50^0$ Celsius and $+50^0$ Celsius to address the outliers. Figure (\ref{fig:thermocouple}) represents the cleaned thermocouple temperatures collected from different trusses. The spikes on the graph reflect the fire intensity during the burn. 
\section{Results }
\label{result}
\subsection{Correlation Analysis}
\begin{figure*}
    \centering
    \begin{subfigure}[b]{0.45\textwidth}
        \centering
        \includegraphics[width=\textwidth]{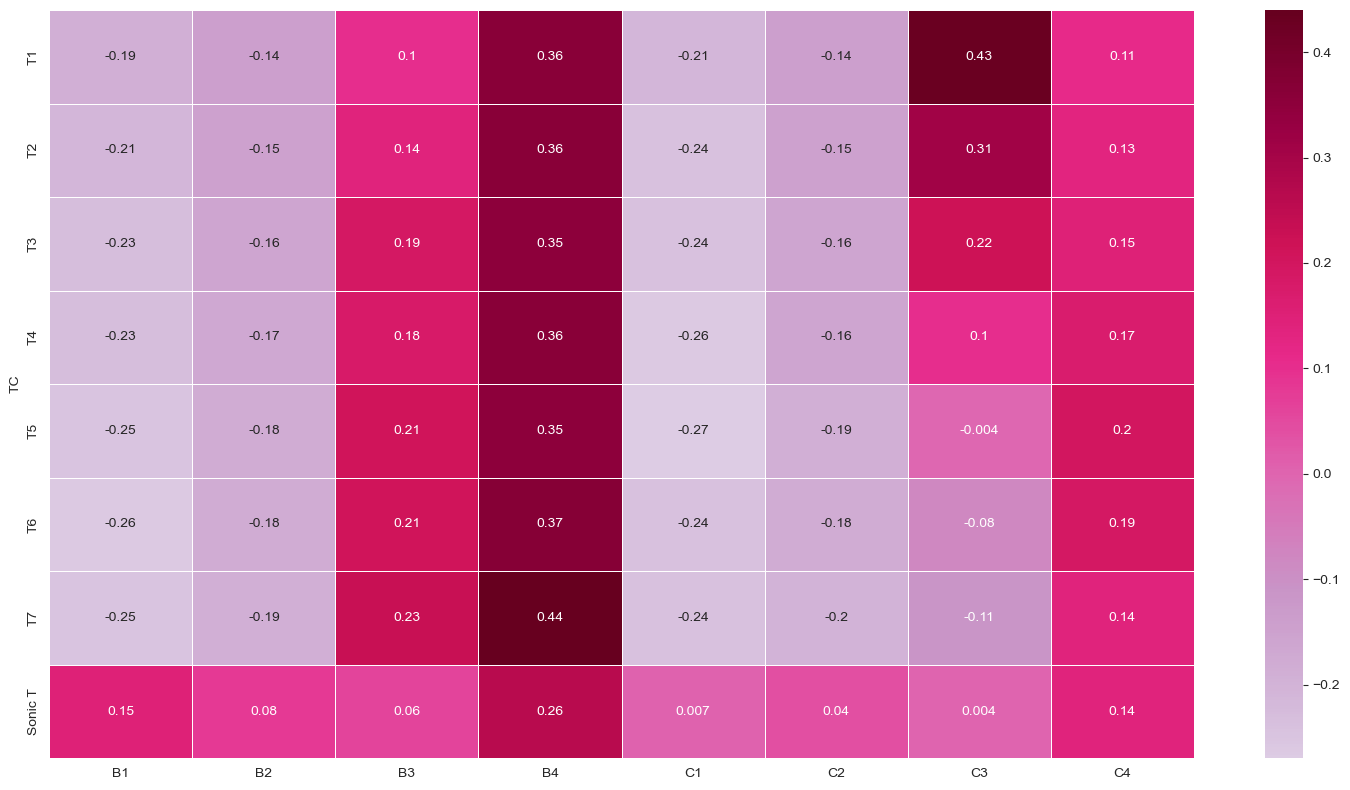}
        \caption{Pearson's Correlation Coefficient }
        \label{fig:pearson}
    \end{subfigure}
    \hfill
    \begin{subfigure}[b]{0.45\textwidth}
        \centering
        \includegraphics[width=\textwidth]{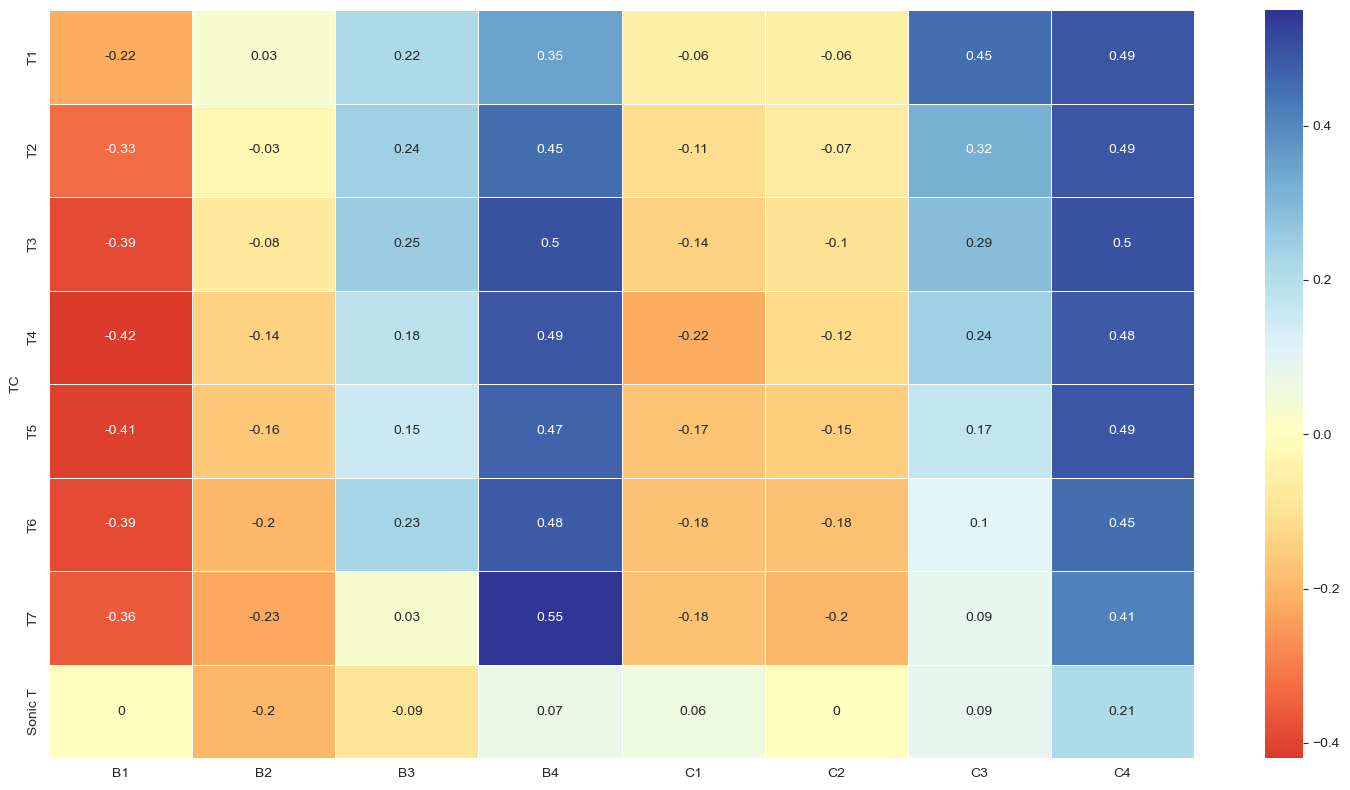}
        \caption{Spearman's Rank Correlation}
        \label{fig:spearman}
    \end{subfigure}
    \caption{Correlation heatmap on datasets from trusses B and C.}
    \label{fig:corr}
\end{figure*}

\begin{figure*}
    \centering
    \begin{subfigure}[b]{0.45\textwidth}
        \centering
        \includegraphics[width=\textwidth]{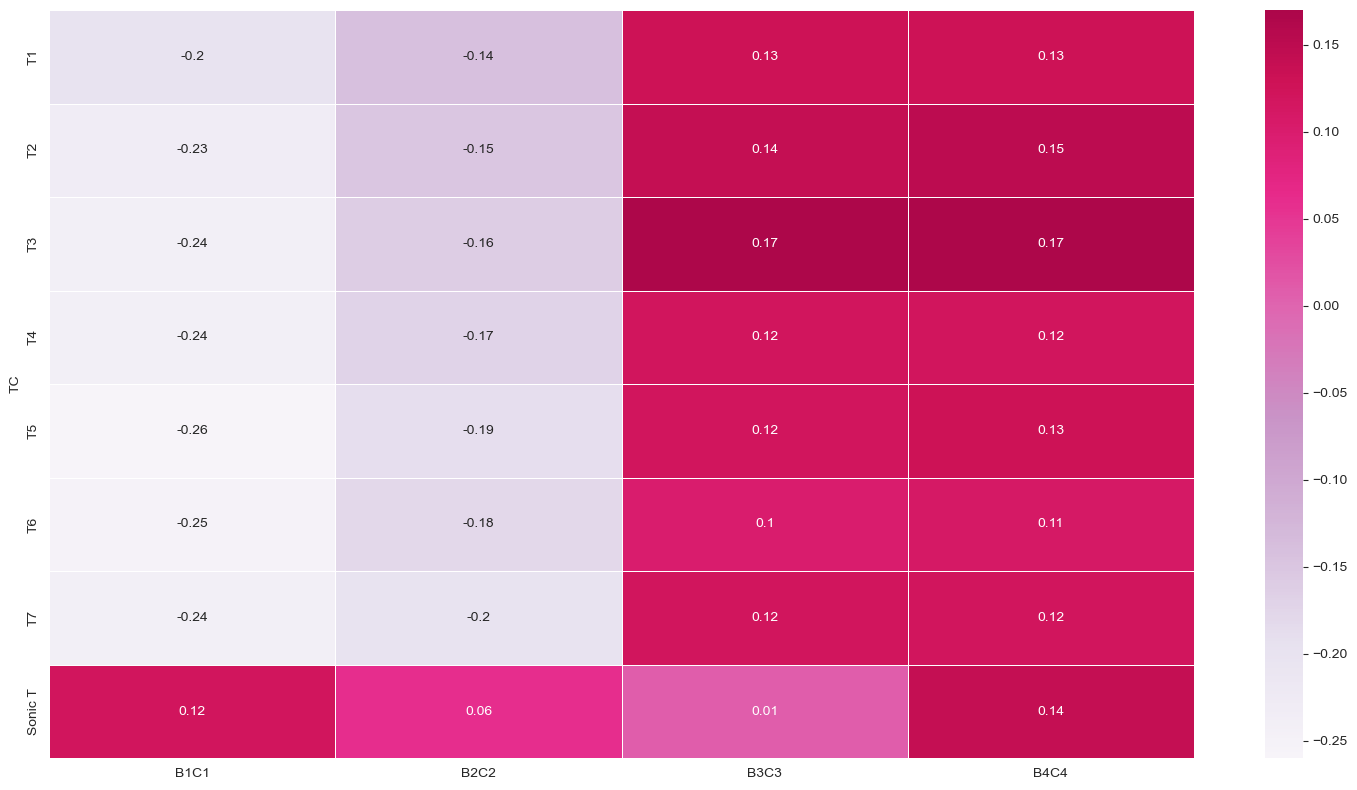}
        \caption{Pearson's Correlation Coefficient }
        \label{fig:pearson2}
    \end{subfigure}
    \hfill
    \begin{subfigure}[b]{0.45\textwidth}
        \centering
        \includegraphics[width=\textwidth]{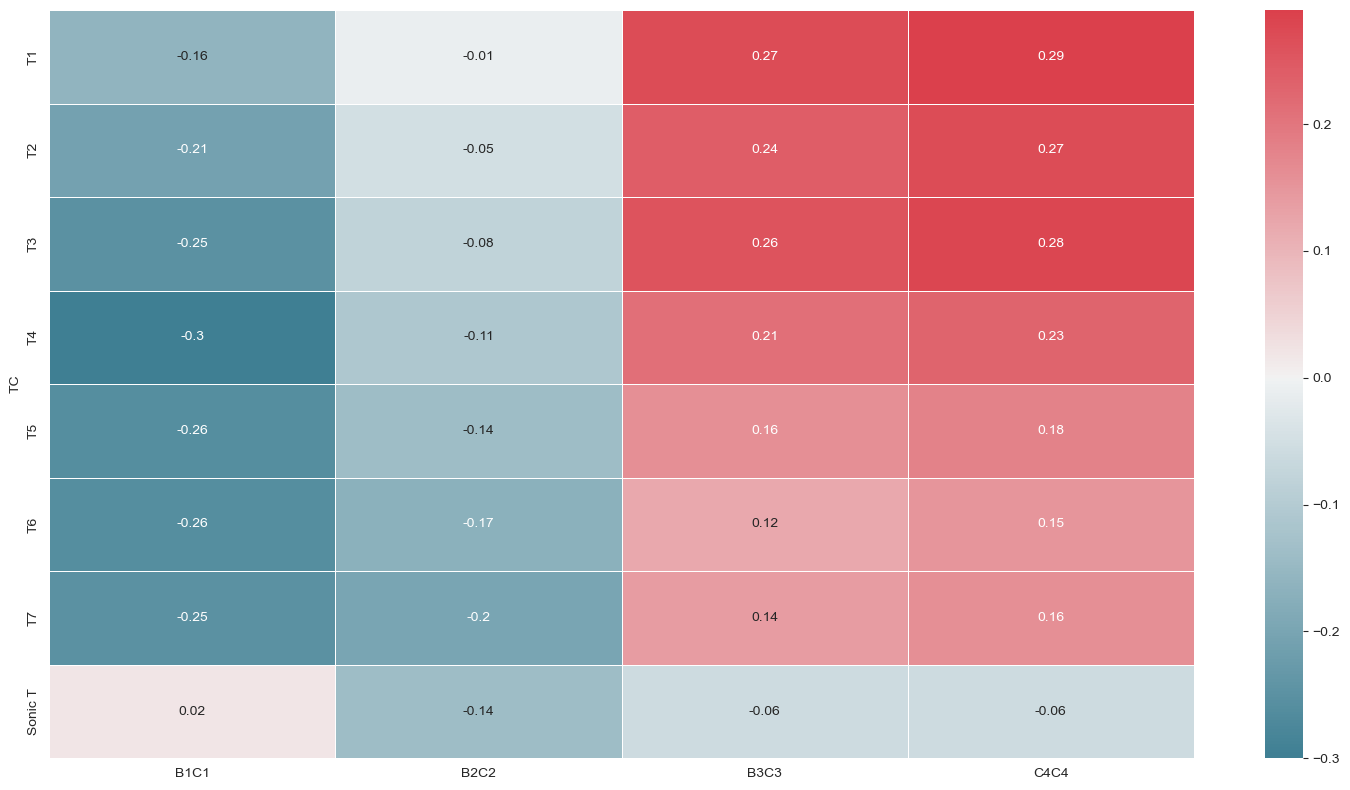}
        \caption{Spearman's Rank Correlation}
        \label{fig:spearman2}
    \end{subfigure}
    \caption{Correlation heatmap on datasets B1C1, B2C2, B3C3,and B3C3}
    \label{fig:corr2}
\end{figure*}

 We calculated  Pearson's Correlation coefficient \citep{pearson} and Spearman's Rank Correlation coefficients \citep{spearman}. These calculations were applied to discern the interrelation between thermocouple temperatures and TKE within each distinct cluster of the dataset. Furthermore, we extended this analysis to combinations derived from pairs of individual clusters. Figures (\ref{fig:pearson}) and (\ref{fig:spearman}) show Pearson's and Spearman's rank correlation coefficients between thermocouple temperatures and TKE for datasets B and C. The correlation for datasets B4 and C4 is significantly stronger and positive, even though the overall coefficients remain subpar. On the other hand, datasets B1 and C1 have the most extreme negative correlation between the two variables. Figures (\ref{fig:pearson2}) and (\ref{fig:spearman2}) demonstrate that combining two sets from four trusses of B and C, B4C4, has the highest positive correlation coefficients.

\subsection{Models Performance Analysis}
Our aim was to estimate TKE using a Deep Neural Network (DNN). We conducted experiments with various numbers of layers and adjusted hyperparameters to obtain the best results. Ultimately, we found that three inner layers and lasso regression regularization in the third layer produced the best results despite having lower accuracy and efficiency than other ensemble models. Detailed evaluations of $R^2$ values of all models over each dataset from the trusses B1- B4 and C1- C4 are presented in the table (\ref{r2_comparison}). Our findings show that DNN models are weaker than other models overall, whereas the Gaussian Processing Regressor (GPR) and Extreme Gradient Boosting (XGB) are the best for each dataset.

\begin{figure*}[!htbp]
    \centering
    \begin{minipage}{\textwidth}
        \centering
        \includegraphics[width = \textwidth]{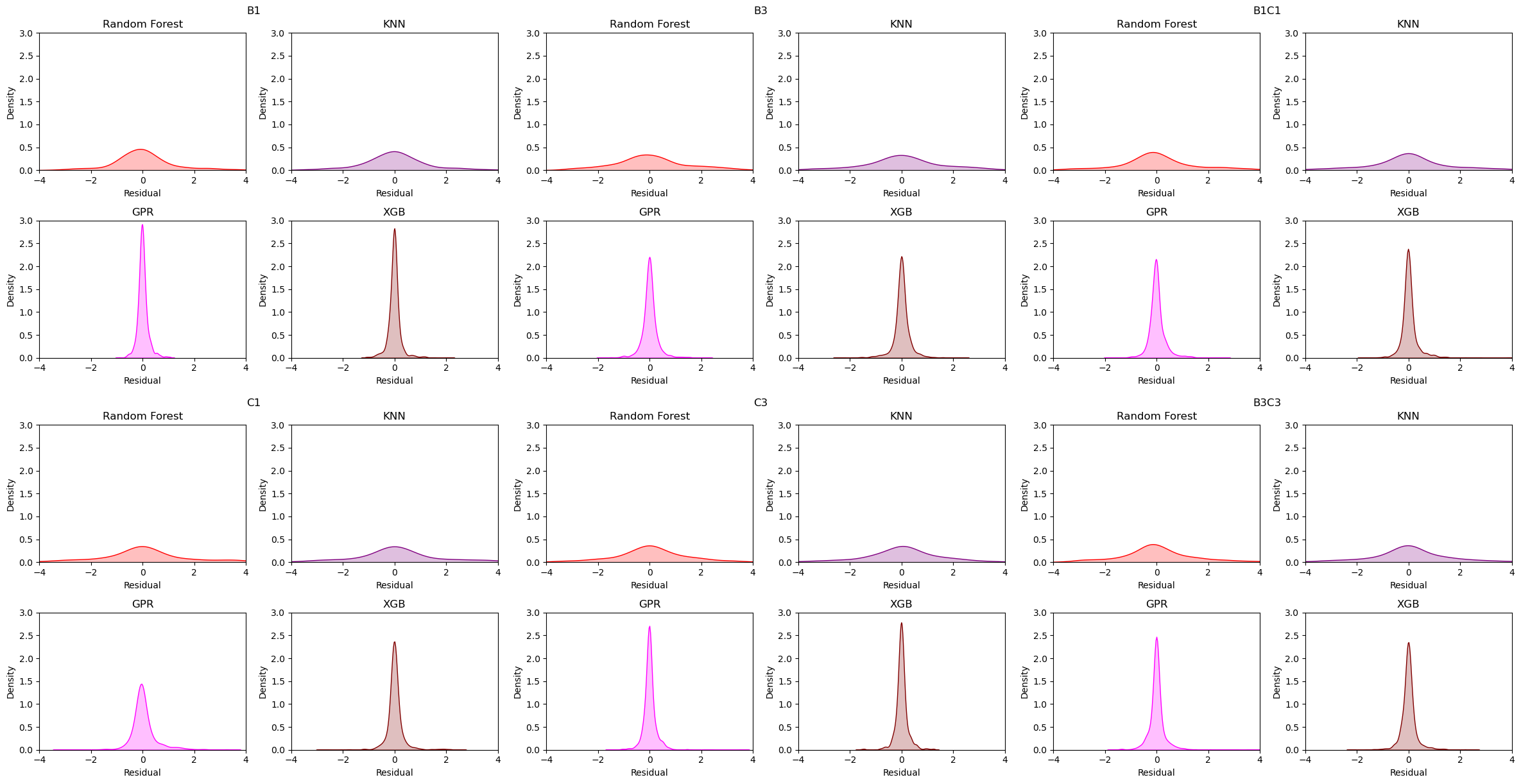}
        \caption{KDE: A residual distribution plot}
        \label{fig:kde}
    \end{minipage}

 \vspace{0.2cm}

    \begin{minipage}{\textwidth}
        \centering
        \includegraphics[width = \textwidth]{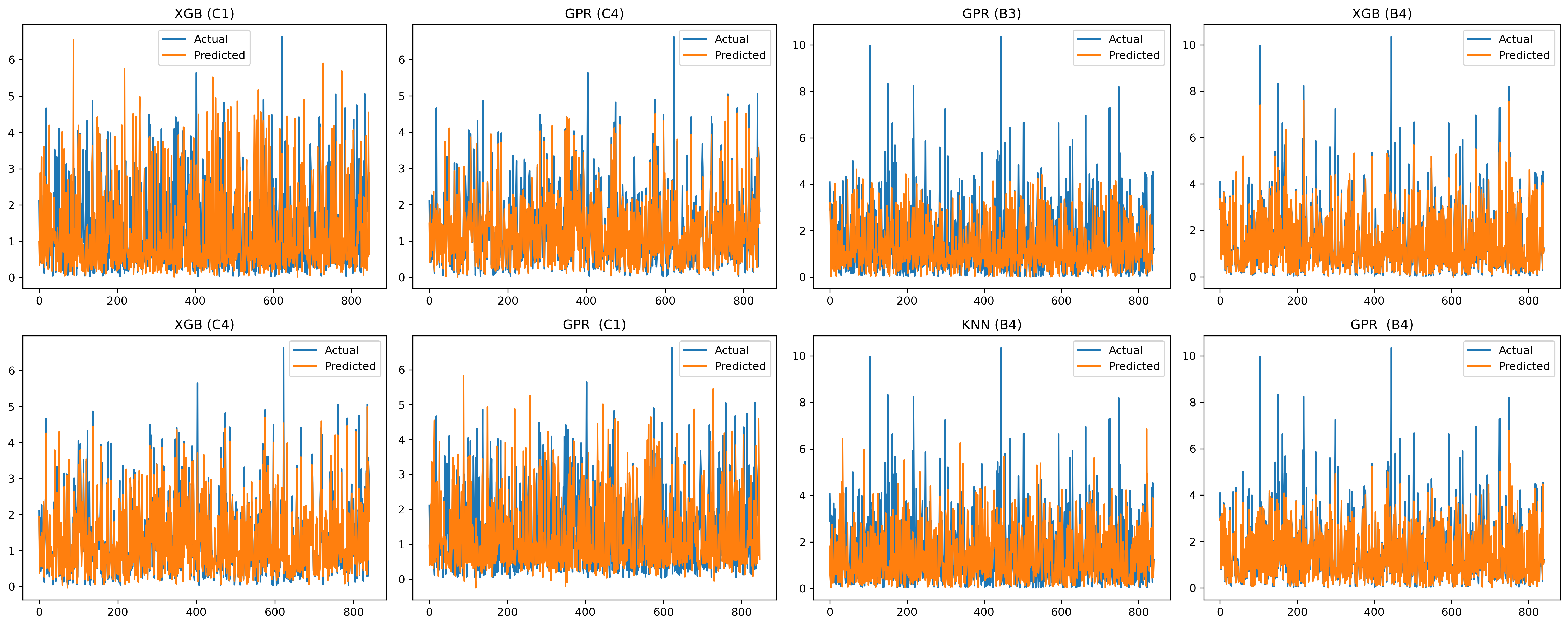}
        \caption{Actual vs predicted values by highly effective models }
        \label{fig:combined}
    \end{minipage}
\end{figure*}

\begin{figure*}[!htbp]
    \centering
    \begin{minipage}{\textwidth}
        \centering
        \includegraphics[width = 0.7 \textwidth]{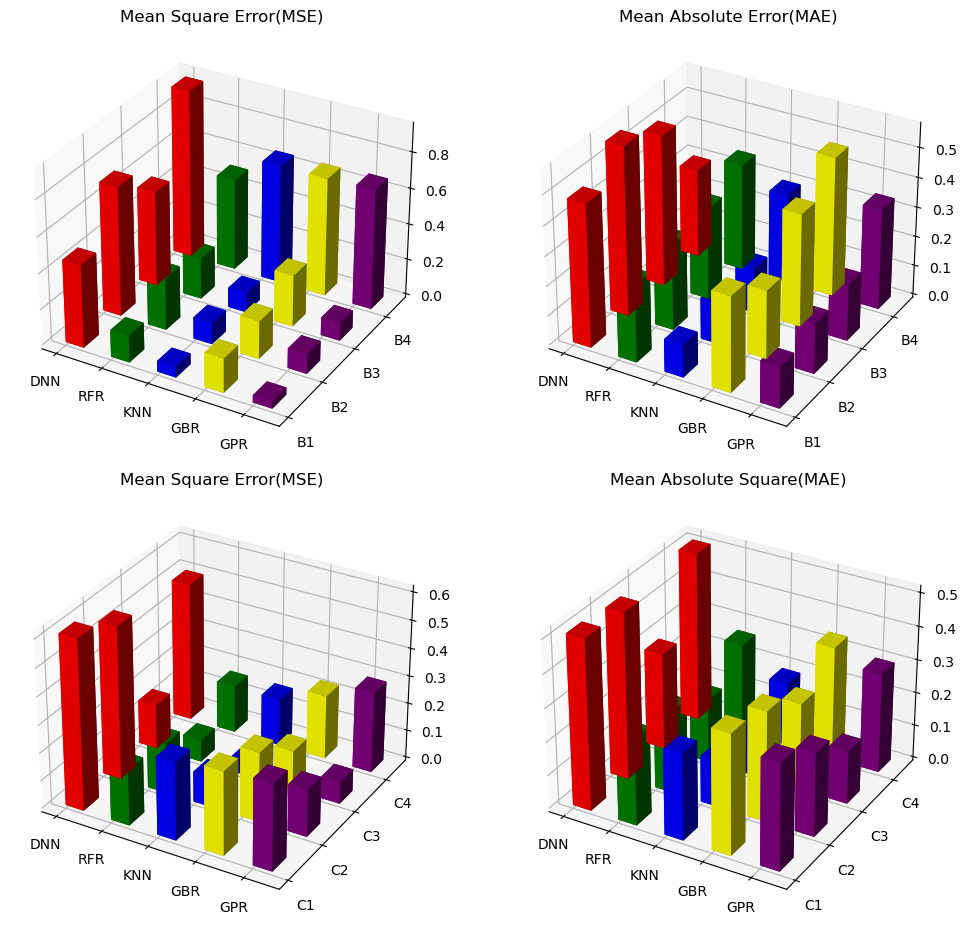}
        \caption{MSE vs MAE comparison over the Models}
        \label{fig:mse-mae}
    \end{minipage}
\end{figure*}

\begin{table*}[!htbp]
        \centering
        \begin{tabular}{@{}l|rrrr|rrrr@{}}
            \toprule
            & \multicolumn{4}{c}{Burn20 B} & \multicolumn{4}{c}{Burn20 C} \\
            \cmidrule(lr){2-5} \cmidrule(lr){6-9}
            ML Models & B1 & B2 & B3 & B4 & C1 & C2 & C3 & C4 \\
            \midrule
            \multicolumn{9}{c}{\textbf{Test (\%)}} \\
            \midrule
            DNN & 52.6 & 52.4 & 60.2 & 64.6 & 60.9 & 61.5 & 63.2 & 61.2 \\
            RFR & 84.8 & 86.0 & 81.2 & 84.5 & 87.3 & 85.6 & 86.6 & 84.3 \\
            KNN & 93.6 & 93.8 & 92.4 & 91.4 & 82.5 & 81.3 & 92.2 & 88.4 \\
            GBR & 79.4 & 80.9 & 75.3 & 79.5 & 81.1 & 81.3 & 82.1 & 81.5 \\
            GPR & 93.7 & 92.6 & 92.4 & 90.4 & 82.4 & 81.3 & 87.2 & 84.4 \\
            XGB & 92.4 & 90.6 & 89.7 & 89.3 & 92.2 & 89.4 & 92.4 & 87.1 \\
            \midrule
            \multicolumn{9}{c}{\textbf{Validation (\%)}} \\
            \midrule
            DNN & 52.6 & 52.4 & 60.2 & 64.6 & 87.7 & 86.7 & 64.8 & 62.2 \\
            RFR & 84.8 & 86.0 & 81.2 & 84.5 & 93.1 & 93.4 & 84.9 & 80.0 \\
            KNN & 93.6 & 93.8 & 92.4 & 91.4 & 96.5 & 95.0 & 84.4 & 82.9 \\
            GBR & 79.4 & 80.9 & 75.3 & 79.5 & 89.0 & 88.4 & 79.8 & 74.4 \\
            GPR & 93.7 & 92.6 & 92.4 & 90.4 & 93.3 & 94.7 & 76.6 & 79.8 \\
            XGB & 92.4 & 90.6 & 89.7 & 89.3 & 95.0 & 94.5 & 89.2 & 87.5 \\
            \bottomrule
        \end{tabular}
        \caption{Comparison of ML Prediction Models over $R^2$}
        \label{r2_comparison}
  
\end{table*}

To evaluate the performance of our models, we analyzed the residual plots and employed Kernel Density Estimation(KDE). KDE is a useful tool for visualizing data distribution over a continuous interval or period \citep{kde}. Our analysis revealed that the GPR model consistently exhibited a sharp peak at the 0 residual for all conditions, indicating superior performance. In contrast, the RF model showed a broader distribution for almost all conditions. See figure (\ref{fig:kde}). \\

The graph shown in figure (\ref{fig:combined}) compares the predicted values of various machine learning models with actual data across four selected datasets. Each subplot displays the actual values in blue and the predicted values in orange. While all models show a decent level of adherence between actual and expected values across datasets, the GPR model has the most consistent results in this expanded dataset, especially in the B3 and B4C4 datasets.\\

The figure (\ref{fig:mse-mae}) presents a performance comparison of six machine learning models across various data conditions using MSE and MAE metrics. GPR, KNN, and XGB consistently showcase superior predictive accuracy across most conditions. In contrast, DNN and RFR occasionally manifest higher errors, suggesting variability in model adaptability to different datasets.

\section{Discussion}

Recent advancements in the quantitative analysis of temperature time series data obtained from thermocouples in experimental fire grids have yielded novel insights into the dynamics of wildland fires. In the past, temperature time series datasets collected from thermocouples within a 10x10m experimental fire grid at a frequency of 10 Hz have been predominantly subjected to qualitative analysis \citep{acp-24-1119-2024,atmos12080956, Observations, acp-24-1119-2024, TKE}, with an emphasis on descriptive trends rather than the extraction of quantitative assessments of underlying physical relationships. However, our research has uncovered a correlation between these datasets and turbulence (TKE) values, highlighting the relationship between temperature fluctuations directly above the fire and nearby turbulence readings. We have developed a methodology to apply this analysis to similar experimental fires and other instruments. Studying the correlation between temperature and TKE has improved our understanding of how temperature changes due to wildfire combustion are linked to turbulence production. This knowledge helps evaluate existing tools and develop new ones for prescribed burns necessary for fuel reduction, forest management, and ecological maintenance.

To rigorously quantify this relationship, we employed spectral analysis and cross-correlation methods, building on turbulence modeling approaches. These methods allowed us to identify periods of heightened turbulence production that corresponded closely with rapid temperature fluctuations, supporting the hypothesis that combustion-driven thermal instabilities are a primary driver of local TKE. Our methodology was validated using the Burn20 dataset, which comprises eight distinct truss-mounted data sets (B1, C1, B2, C2, B3, C3, B4, C4) and their combinations (B1C1, B2C2, B3C3, B4C4), collected at varying spatial coordinates and heights. The analysis revealed that the correlation between temperature and TKE is not spatially uniform; rather, it is highly dependent on the position and elevation of the measurement trusses. This spatial heterogeneity aligns with findings in \citep{finney}, where local topography and fuel distribution were shown to influence fire-induced turbulence patterns.

 Furthermore, the study revealed a significant connection between temperature and TKE time series, shedding light on how temperature fluctuations caused by combustion influence turbulence production above and near wildland fires. However, we observed that the correlation between temperatures and TKE varies across different trusses, with certain trusses contributing more significantly than others. These findings provide valuable insights for wildland fire researchers, enabling them to evaluate existing tools and develop new strategies for fire management. Such advancements will aid fire managers in planning and executing prescribed burns, essential for fuel reduction, forest management, and the restoration and maintenance of ecological balance.

Artificial Intelligence and Machine Learning algorithms have been employed in the domain of wildfire science and management, utilizing a diverse range of wildfire datasets \citep{doi:10.1139/er-2020-0019}. A significant focus has been placed on predictive frameworks, such as Recurrent Neural Networks (RNNs), Convolutional Neural Networks (CNNs), and various ensemble regression models, which are applied to remote sensing data pertaining to wildfires \citep{Andrianarivoly, machine_Remote1, machine_Remote2, machine, BJANES, MAJUMDER2025102956, Shadrin2024}.Certain researchers have employed Reinforcement and Supervised Machine Learning techniques, along with their combination in Long-Term Recurrent Convolutional Networks (LRCH), using Sentinel-1 Ground Range Detected data \citep{reinforcement, reinforcement2,unsupervised}.  Following an exhaustive correlation analysis, for the first time, we developed and explored a comprehensive array of machine learning ensemble regression models and deep neural network architectures to investigate the impact of temperature variations during combustion events on fire turbulence. This endeavor successfully facilitated the quantification of fire turbulence through Turbulent Kinetic Energy (TKE) values, determined from thermocouple temperature readings. Leveraging these temperature fluctuations permits an effective estimation of the TKE associated with wildland fires, thereby revealing critical patterns that influence fire dynamics. By applying deep neural networks alongside a diverse array of sophisticated ensemble regression models to high-frequency (10Hz) temperature and turbulence data derived from small-scale experimental prescribed burns, we have significantly advanced our capacity to quantify turbulence and fire propagation behavior. Amongst the ensemble methodologies evaluated, the Gaussian Process Regressor and Extreme Gradient Boosting algorithms demonstrated exceptional proficiency and precision in predicting TKE values. This capability not only enhances the quantification of fire turbulence but also facilitates a deeper understanding of the underlying mechanisms governing fire behavior. The integration of these cutting-edge techniques paves the way for improved predictive modeling of wildland fires, specifically with diverse temperature fluctuation, which contributes to planning early, to better resource management, and enhanced safety measures in fire-prone areas.

\section{Conclusion}
\label{sec:conclusions}

We investigated the relationships between thermocouple temperatures, sonic temperatures, and Turbulent Kinetic Energy (TKE).
While the correlation and covariance between the predictor variables and the target variable were not particularly strong, our machine learning and deep learning algorithms successfully uncovered hidden and intricate patterns within the predictor variables relevant to the target variable.
As a result, we achieved accurate fire behavior predictions and spread through TKE estimation using these advanced algorithms. Through extensive hyperparameter tuning, exploration, and the implementation of complex regularization techniques, we identified Gaussian Process Regressors as the most effective model for estimating TKE with high precision and minimal error.

In our subsequent research, we aim to further explore fire behavior and propagation by integrating high-dimensional infrared (IR) image datasets with the complex physical dynamics of associated tools. A key focus will be leveraging advanced Deep Neural Network (DNN) methodologies to analyze hyperspectral image datasets related to wildland fires. Additionally, we plan to investigate the role of atmospheric conditions and fuel moisture content in modulating fire dynamics, as these factors are critical for improving the accuracy of fire spread models. By combining data-driven approaches with physical modeling, we hope to develop more robust tools for predicting and mitigating the impacts of wildland fires in diverse environments.

Furthermore, we aim to investigate the potential of real-time data assimilation techniques to improve the predictive capabilities of our models. This will involve integrating live sensor data with our existing frameworks to provide timely and accurate forecasts of fire behavior. Such innovations will advance scientific understanding and provide actionable insights for fire management teams in the field. Additionally, we will investigate the impact of terrain and vegetation heterogeneity on fire spread, as these factors play a crucial role in determining the intensity and direction of wildfires. By incorporating these elements into our models, we aim to create a more comprehensive framework for wildfire prediction and management.\\

\section*{Funding}
This research was funded by the National Science Foundation Oak Ridge Institute of Science and Education (NSF-ORISE) under the Mathematical Science Graduate Internship (MSGI). Data acquisition was supported by the US Department of Defense Strategic Environmental Research and Development Program Project RC-2641.CN is partially funded by NSF MCB 2126374.
\section*{Data Availability}
All data is publicly available and cited. Code available upon request.
\section*{CRediT authorship contribution statement}
\textbf{Dipak Dulal:} Writing original draft, Methodology, Formal analysis, programming, conceptualization. \textbf{J. J. Charney:} Writing-review and editing, Supervision, Methodology, Conceptualization, interpretation, and data provision. \textbf{M. Gallagher:} Writing-review and editing, data preparation and interpretation, Supervision. \textbf{P. Acharya:} Writing-review \& editing, methodology, Formal analysis, 
numerical interpretation. \textbf{C. Navasca:} Writing-review \& editing, Supervision. \textbf{N. Skowronski:} Writing-review\& editing, Supervision, Methodology, funding acquisition, data acquisition, Conceptualization.
\section*{Declaration of completing interest}
The authors hereby declare that they have no known competing financial interests or personal relationships that could have potentially influenced the work presented in this paper. 

\bibliographystyle{elsarticle-harv}    
\bibliography{usfs.bib}

\end{document}